\documentclass[runningheads]{llncs}
\usepackage[vietnamese,english]{babel}
\usepackage{multirow}
\usepackage{graphicx}
\usepackage{caption}
\usepackage{subcaption}
\usepackage{amsmath}
\begin{document}
\title{Improving Vietnamese Legal Question--Answering System based on Automatic Data Enrichment}

\titlerunning{Improving Vietnamese Legal QAS based on Automatic Data Enrichment}

\author{Thi-Hai-Yen Vuong\inst{1} \and
Ha-Thanh Nguyen\inst{2} \and
Quang-Huy Nguyen\inst{1} \and \\
Le-Minh Nguyen\inst{3} \and
Xuan-Hieu Phan\inst{1}
}
\authorrunning{Vuong et al.}
%
\institute{VNU University of Engineering and Technology, Hanoi, Vietnam \email{\{yenvth,19020011,hieupx\}@vnu.edu.vn} \and
National Institute of Informatics, Tokyo, Japan \email{nguyenhathanh@nii.ac.jp} \and
Japan Advanced Institute of Science and Technology, Ishikawa, Japan \email{nguyenml@jaist.ac.jp}
}

\maketitle              
\begin{abstract}
  Question answering (QA) in law is a challenging problem because legal documents are much more complicated than normal texts in terms of terminology, structure, and temporal and logical relationships. It is even more difficult to perform legal QA for low-resource languages like Vietnamese where labeled data are rare and pre-trained language models are still limited. In this paper, we try to overcome these limitations by implementing a Vietnamese article-level retrieval-based legal QA system and introduce a novel method to improve the performance of language models by improving data quality through weak labeling. Our hypothesis is that in contexts where labeled data are limited, efficient data enrichment can help increase overall performance. Our experiments are designed to test multiple aspects, which demonstrate the effectiveness of the proposed technique.

\end{abstract}
\section{Introduction}
The performance of question-answering (QA) has increased significantly thanks to the rapid development and recent breakthroughs in natural language processing. With these advances, QA has been used actively in various business domains in order to save human labor, get more automation as well as enhance user experience. Among application areas, QA in the legal domain has attracted a lot of interest from the research community as well as the awareness and support from legal practitioners, experts, law firms, and government agencies. Legal QA could assist them to find relevant legal information quickly, accurately, and reliably.

Technically, the legal retrieval-based QA problem is simply stated as follows: given a query $q$ and a text corpus $D = \{d_1, d_2, \dots, d_n\}$, 
the retrieval-based QA finds the most likely document $d^*$ that maximizes the relevance score $R$:

\begin{equation}
d^* = \arg\max_{d \in D} R(q,d)
\end{equation}

\noindent where $R(q,d)$ represents the relevance score of the query $q$ and document $d$.

Traditionally, lexical weighting and ranking approaches like TF-IDF or BM25 are used to find the relevant documents based on the match of vocabulary terms. Despite their limited accuracy and simplicity, these techniques are normally cost-effective. Meanwhile, representation and deep learning based models are likely to give better results but they are much more expensive in terms of large training data, computing power, storage, and deployment. Various deep learning models have been introduced to enhance the representation of queries and documents, such as CNN \cite{hu2014convolutional}, RNN and LSTM \cite{palangi2016deep,tai-etal-2015-improved}. Pre-trained language models (BERT \cite{devlin2018bert}, GPTs \cite{brown2020language}) also significantly improve text representation in retrieval tasks. 

In the legal domain, there are several challenges to building a reliable QA system. First, legal documents are much more complex than normal texts. They contain legal terms and concepts that are not commonly observed in general texts. Legal texts are usually long and have complex structures. There are also temporal constraints, logical relations, cross--document references etc. that are even difficult for human readers to follow and understand. Second, data annotation for legal documents is a real challenge, making it hard to construct even a medium-sized high-quality labeled dataset for training QA models.

Today, one popular way to improve accuracy is to build large deep-learning models with a huge number of parameters. This is obviously an obstacle because building such models requires powerful computing resources and a huge source of data. In this work, we want to concentrate on enhancing data quality and quantity in the context where expanding labeled data is infeasible. A heuristic method for automatically creating weak label datasets and supporting relationship representation models in case law retrieval is presented by Vuong et al. \cite{vuong2022sm}. Therefore, we apply this technique to create more training data to improve our models without the need of increasing number of model parameters.

Technically, we address the problem of article-level retrieval-based legal QA. We use the Vietnamese civil law QA dataset, which was introduced by Nguyen et al. \cite{nguyen2022attentive}, to conduct an empirical study on the proposed methods. Table \ref{tab:example} illustrates an example of a legal query and the anticipated response. It is difficult to represent, retrieve and determine the correct answer when the articles are often long and complex. In addition, a notable feature of this dataset is that each article usually has a title, which serves as a brief summary.

\begin{table*}[ht]
\caption{A sample in the dataset}
\selectlanguage{vietnamese}
\label{tab:example}
\begin{tabular}{|p{0.15\linewidth} | p{0.85\linewidth}|}
\hline
\textbf{Question} & Hợp đồng ủy quyền có hiệu lực khi đáp ứng tiêu chí nào? \\ 
 & \textit{(An authorization contract is effective when it meets what criteria?)} \\ \hline
\textbf{Answer} & Article 117 form Document 91/2015/QH13 \\ \hline
\textbf{Article} & Điều kiện có hiệu lực của giao dịch dân sự \\ 
\textbf{Title} & \textit{(Valid conditions of civil transactions)} \\ \hline
\textbf{Article} & Giao dịch dân sự có hiệu lực khi có đủ các điều kiện sau đây: \\ 
Content & a) Chủ thể có năng lực pháp luật dân sự, năng lực hành vi dân sự phù hợp với giao dịch dân sự được xác lập; \\ 
 & b) Chủ thể tham gia giao dịch dân sự hoàn toàn tự nguyện; \\ 
 & c) Mục đích và nội dung của giao dịch dân sự không vi phạm điều cấm của luật, không trái đạo đức xã hội. Hình thức của giao dịch dân sự là điều kiện có hiệu lực của giao dịch dân sự trong trường hợp luật có quy định. \\ 
 & \textit{(A civil transaction takes effect when the following conditions are satisfied:} \\ 
 & \textit{a) The subject has civil legal capacity and civil act capacity suitable to the established civil transactions;} \\ 
 & \textit{b) Entities participating in civil transactions completely voluntarily;} \\ 
 & \textit{c) The purpose and content of the civil transaction do not violate the prohibition of the law and do not violate social ethics. The form of a civil transaction is the effective condition of a civil transaction in case it is provided for by law.)} \\ \hline
\end{tabular}%
\end{table*}

The main contributions of our work are twofold. First, we built an end-to-end article retrieval system to solve the legal QA task. Second, we show how efficient automated data enrichment is and we conducted a variety of experiments to contrast our model with the most cutting-edge approaches in this domain. 

\section{Related Work}
In natural language processing, the term question answering (QA) is commonly used to describe systems and models that are capable of providing information based on a given question. Depending on the characteristics of the task, we can divide it into different categories. Factoid QA \cite{iyyer2014neural} is a class of problems for which the answer is usually simple and can be further extracted from a given question or context. Problems in this category can often be solved with generation models or sequence tagging approaches. Retrieval-based QA \cite{feldman2019multi} is a class of problems where the answer should be retrieved from a large list of candidates based on relevancy and ability to answer the question. This class of problems can also be called List QA. Confirmation QA \cite{sanagavarapu2022disentangling} is the class of problems where systems or models need to confirm whether a statement is true or false. Systems for this type of problem can be an end-2-end deep learning model, knowledge-based systems, or neuro-symbolic systems.

In the legal field, question-answering has been posed in the research community for many years \cite{rabelo2022overview}. The main challenges of this problem on the rule language include fragmented training data, complex language, and long text.
With the emergence of transformer-based \cite{vaswani2017attention} language models as well as transfer learning and data representation techniques, the performance of systems on tasks is significantly improved.
In legal information retrieval, a number of neural approaches are also introduced to address the problem of word differences and characteristics of complex relationships \cite{huang2013learning,sugathadasa2019legal,tran2020encoded,nguyen2022attentive}.

\section{Dataset}


\textbf{Original dataset}: the corpus is collected from Vietnamese civil law. The labeled dataset was introduced by Nguyen et al. \cite{nguyen2022attentive}. Table \ref{tab:analys1} \& \ref{tab:analys2} give a statistical summary of the corpus and dataset. There are 8587 documents in the corpus. Vietnamese civil law documents have a long and intricate structure. The longest document contains up to 689 articles, and the average number of articles per document is also comparatively high at 13.69. The average title length in this dataset is 13.28 words, whereas the average content length is 281.83 words. 

\begin{table}
\caption{Corpus of Vietnamese legal documents statistics}
\label{tab:analys1}
\centering
\begin{tabular}{lc}
\hline
\textbf{Attribute} & \textbf{Value} \\ \hline
Number of legal documents & 8,587 \\
Number of legal articles & 117,557 \\
Number of articles missing title & 1,895 \\
The average number of articles per document & 13.69 \\
Maximum number of articles per document & 689 \\
The average length of article title & 13.28 \\
The average length of article content & 281.83 \\
\hline
\end{tabular}%
\end{table}

This is also worth noting  because one of the challenges and restrictions is the presentation of long texts. On average, the questions are less than 40 words long. Because of the similarity in their distributions, it is expected that the model trained on the training set will yield good performance on the test set.

\begin{table}
\caption{Original dataset statistics}
\label{tab:analys2}
\centering
\begin{tabular}{lcc}
\hline
 & \textbf{Train set} & \textbf{Test set} \\ \hline
Number of samples & 5329 & 593 \\ 
Minimum length of question & 4 & 5 \\ 
Maximum length of question & 45 & 43 \\ 
Average length of question & 17.33 & 17.10 \\ 
Minimum number of articles per query & 1 & 1 \\ 
Maximum number of articles per query & 11 & 9 \\ 
Average number of articles per query & 1.58 & 1.60 \\ \hline
\end{tabular}%
\end{table}

\textbf{Weak labeled dataset}: Vuong et al. have the assumption that the sentences in a legal article will support a topic sentence \cite{vuong2022sm}. On the basis of this supposition, the weak labeled dataset is created. There is also a similar relationship in this dataset. The title serves as a brief summary of the article, so the sentences in the article content support to title. We apply this assumption to our method. By considering the title to be the same as the question, we will produce a dataset with weak labels. A title and content pair would be a positive example equivalent to a question and related articles pair. We randomly generated negative examples at a ratio of 1:4 to positive labels and obtained a weak label dataset consisting of 551,225 examples.

\section{Methods}

For a legal question-answering system at the article level, given a question $q$, and a corpus of Civil Law $CL = \{D_1,D_2,...,D_n\}$, the system should return a list of related articles $A = \{a_i|a_i \in D_j, D_j \in CL\}$. The following section provides a detailed description of the phases involved in resolving the problem.

\subsection{General Architecture}

\begin{figure*}[ht]
    \centering
    \includegraphics[width=\textwidth]{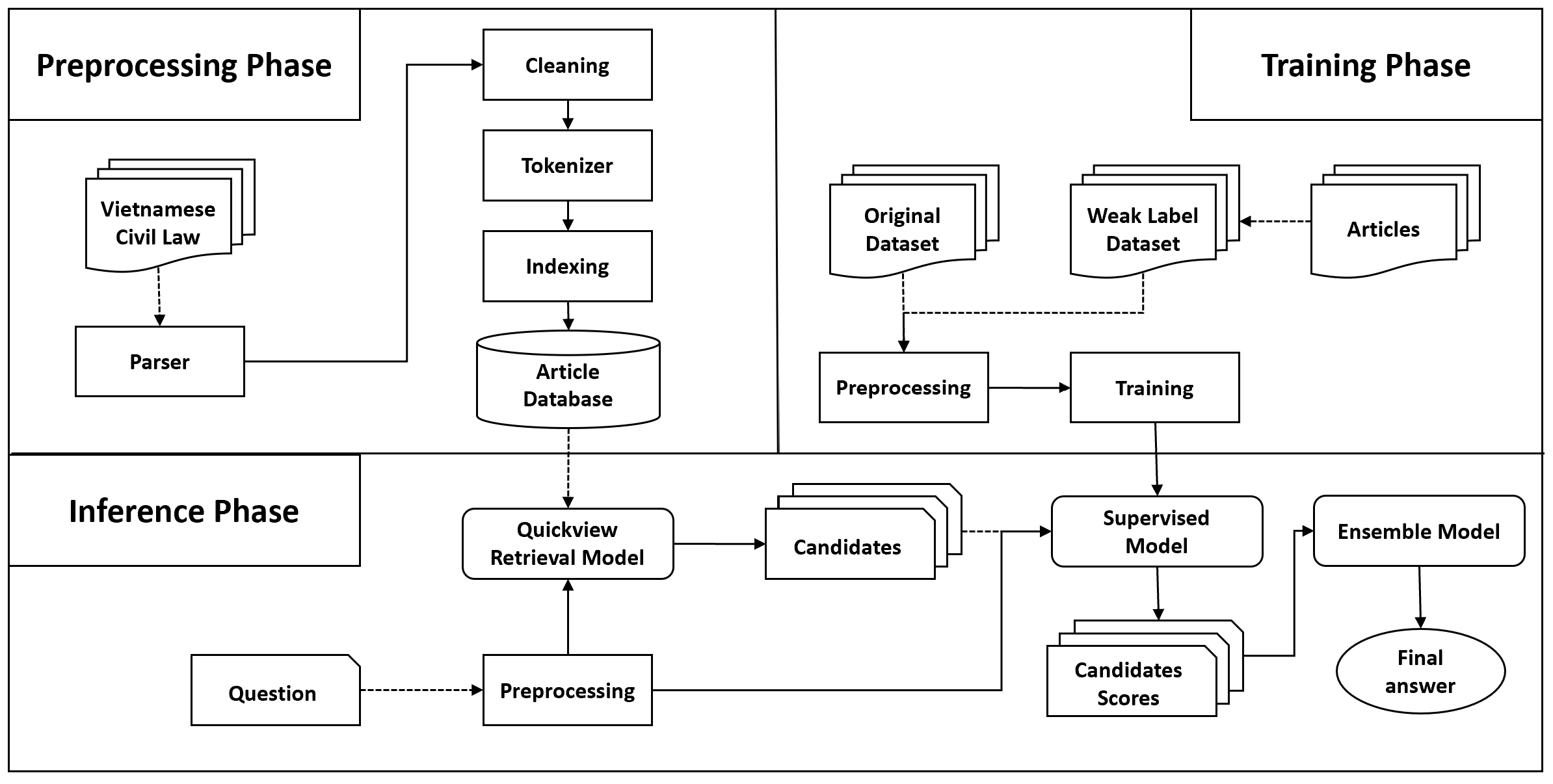}
    \caption{Pipeline in the end-to-end article retrieval-base question answering system}
    \label{fig:pipeline}
\end{figure*}

Figure \ref{fig:pipeline} demonstrates our proposed system. There are three main phases: preprocessing, training, and inference phase.

\textbf{Preprocessing phase}: the result of this phase is an article-level database, which involves processing the raw Vietnamese civil law documents.

\begin{itemize}
    \item \textbf{\textit{Vietnamese Civil law}} is a corpus of Vietnamese legal documents.
    \item \textbf{\textit{Parser}} segment legal documents into list of articles.
    \item \textbf{\textit{Cleaning}} will filter out documents with metadata. Special symbol characters are also removed from the article. Numbers and vocabulary are retained and converted to lowercase.
    \item \textbf{\textit{Tokenizer}} is crucial to the processing of Vietnamese natural language. Vietnamese word structure is quite complicated, a word might contain one or more tokens. 
    \item \textbf{\textit{Indexing}} is a task to represent and put articles into the database. Given a query, the search engine will return the response quickly and accurately.
\end{itemize}

\textbf{Training phase}: we construct a supervised machine-learning model to rank the articles pertaining to the input question.

\begin{itemize}
    \item \textbf{\textit{Original dataset}} is a legal QA dataset provided by Nguyen et al. \cite{nguyen2022attentive}.
    \item \textbf{\textit{Articles}} is result of the preprocessing phase.
    \item \textbf{\textit{Weak label dataset}} was create by our heuristic method.
    \item \textbf{\textit{Preprocessing}} includes tasks similar to the preprocessing phase for question processing.
    \item \textbf{\textit{Training}}, we will construct a deep learning model to rank the texts related to the question. 
\end{itemize}

\textbf{Inference phase}: is the process to generate the response to a new input question.
\begin{itemize}
    \item \textbf{\textit{Question}} is query in natural language. 
    \item \textbf{\textit{Preprocessing}} is same as previous phases to process input question.
    \item \textbf{\textit{Quickview retrieval model}} matches questions and texts using unsupervised machine learning techniques . The processing speed of this model is typically fast.
    \item \textbf{\textit{Candidates}} are a list of limited candidates returned from quickview retrieval model.
    \item \textbf{\textit{Supervised model}} is result of the training phase. Its inputs are the question and the article candidates.
    \item \textbf{\textit{Candicate scores}} are outputs of Supervised model.
    \item \textbf{\textit{Ensemble model}} will combine the scores of the quickview retrieval model and the supervised model to make a final decision.
\end{itemize}

\subsection{Indexing}

There are numerous methods for indexing text into a database; in this work, we conduct experiments in two ways: word indexing and dense indexing.

\textbf{Word indexing}: During the indexing process, the words in the text will be analyzed, normalized, and assigned a corresponding index. When given a query, the system searches the index the most related. Word indexing helps to find and look up information in the text faster and more accurately.

\textbf{Dense vector indexing}: In addition to word indexing, word-to-vec and sequence-to-vec are both common methods for representing text semantically. These dense vectors can be used to represent text and index the database for search purposes. We apply two ways of representing text as dense vector according to w2v (FastText \cite{joulin2017bag}) and contextual embedding (BERT \cite{devlin2018bert}) to encode the given question and the legal articles. FastText is a model that converts each word into a dense vector of 300 dimensions. To construct a vector representation of a text, we average over the word vectors to form a single representation vector. Sentence-BERT converte the text into a dense vector with 768 dimensions that can represent the contextual semantics of the document by the Sentence-BERT model \cite{reimers2019sentence}. 

Table \ref{tab:analys1} shows that the length of articels is often large, which is a limitation of the text representation by FastText and BERT. On the other hand, most questions just partially match articles, we overcome this long presentation weakness by splitting the legal article into a list of sentences and then generating dense vectors before indexing them into the database.

\subsection{Quickview Retrieval Model}

There are 117,575 legal articles in this corpus. This is a huge number, so in order to ensure the effectiveness of the question-answering system, we build a so-called quickview retrieval model using unsupervised machine learning techniques in order to rapidly return a limited candidate set.

\textbf{Word matching}: to compare questions and articles in the word indexing database, we use the BM25 algorithm \cite{robertson1994some}. The bag-of-words retrieval function BM25 estimates the relevance of a document to a given search query by ranking documents according to the query terms that appear in each document.

Given a question $Q$, containing tokens $\{t_{1},t_{2},..., t_{n}\}$, the BM25 score of a article $A$ is:
\begin{equation}
    BM25S(Q,A) = \sum_{i=1}^n IDF(t_{i}) \cdot \frac{f(t_{i},A) \cdot (k_{1}+1)}{f(t_{i},A) + k_{1} \cdot (1-b+b \cdot \frac{|A|}{avgdl})}
\end{equation}
in which:
\begin{itemize}
    \item $f(t_{i},A)$: $t_{i}$'s term frequency in the legal article $A$
    \item $|A|$: a number of word in in the legal article $A$ in terms
    \item $avgdl$: the average article length in the legal corpus. 
    \item $k_{1}$: a saturation curve parameter of term frequency. 
    \item $b$: the importance of document length. 
    \item $IDF(t_{i})$ is the inverse document frequency weight of the given question $t_{i}$, follow as: $IDF(t_{i}) = \ln(1 + \frac{N - n(t_{i} + 0.5)}{n(t_{i}) + 0.5})$. $N$ is amount of articles in the legal corpus, and $n(q_{i})$ is amount of articles containing $q_{i}$.
\end{itemize}

While a content article is intense with full meaning, the article title contains a significant meaning. In this instance, the quickview retrieval score is determined using the formula below:

\begin{equation}
    QS(Q,A) = \alpha * BM25S(Q,TA) + \beta * BM25S(Q,CA)
\end{equation}
in which, $\alpha$ and $\beta$ are boosting weights. $TA$ and $CA$ are the titles and content of the article.

\textbf{Dense vector matching}: to estimate the semantic similarity between questions and legal articles in the dense indexing database, we use cosine similarity to calculate quickview retrieval score:

\begin{align}
    Cosine(VQ,VSA) &= \frac{VQ^T \cdot VSA}{||VQ|| \cdot ||VSA||} \\
    QS(Q,A) &= \underset{1 \leq j \leq n}\max(Cosine(VQ,VSA_j))
\end{align}

in which, $VQ$ is presentation vectors of the question.0. $VSA_j$ is presentation vectors of the $j^{th}$ sentence in the legal article. $n$ is the number of sentences in the legal article. Finally, We use minmaxscaler to normalize scores and generate a list of ranked candidates.

\subsection{Supervised Model}

Pre-trained language models have proven useful for natural language processing tasks. Particularly, BERT significantly enhanced common language representation \cite{devlin2018bert}. We use the BERT pre-training model and adjust all its parameters to build the related classifier model. We use the first token's final hidden state $h$ as the presentation for the question-article pair. The last layer is a single fully connected added on the top of BERT. The output of the model is a binary classification. Cross-entropy loss is applied to the loss function. Adam \cite{adam} is used to optimize all model parameters during the training phase with a learning rate of $e^{-5}$. The supervised score between the question and the legal article is the classification probability of label 1:

\begin{equation}
    SS(Q,A) = P_{label=1}(Q,A)
\end{equation}

Lastly, we also use minmaxscaler to normalize scores and reranking a list of candidates. In this model, we proceed to build a related classification model based on two training datasets: the original dataset and a full dataset (original and weak label dataset). In the training process with the full dataset, we fit the model on weak label data first. Then use the best model to fine-tune with the original dataset.

\subsection{Ensemble Model}

We utilize the quickview retrieval model to generate a list of the $top-k$ candidates. These candidates are then refined using a supervised ensemble model, which provides higher precision but is slower. The quickview model serves as a preliminary selection step due to its fast computation despite its lower precision.

We use a variety of measures of similarity, including lexical similarity (the quickview retrieval model) and semantic similarity (the supervised model). Despite the fact that lexical and semantic similarities are very different from one another, they can work in tandem and are complementary. The combined score of the question $Q$ and the candidate article $CA_i$ is calculated as follows:

\begin{equation}
    CombineS(Q,CA_i) = \gamma * QS(Q,CA_i)
    + (1-\gamma) * SS(Q,CA_i)
\end{equation}
where $\gamma \in [0,1]$.

Table \ref{tab:analys1} indicates that each question can have one or more related articles (the average is about 1.6). The most relevant article $MRCA$ is returned by default, to determine a set of candidates to return, we would normalize the combined score and use the threshold parameter: a final returned articles set $FRA = \{CA_i|CombineS(Q,MRCA) - CombineS(Q,CA_i)$ $< threshold\}$.

\section{Experimental Results and Discussion}
To ensure fairness in the training process and selection of hyperparameters, we divided the training dataset into training and validation with a ratio of 9:1.  
In the quickview retrieval phase, we utilize the $Recall@k$ measure to assess the list of returned candidates. Recall@k is (Number of correctly predicted articles in the $top-k$ results) / (Total number of gold articles). Macro-F2 is a metric to evaluate the end-to-end question-answering system. Precision, recall, and average response time per question are also used to evaluate the system's performance.

The processing phase and the quickview retrieval model are carried out on CPU Intel core i5 10500 and 32Gb ram. The supervised model is trained and inference on NVIDIA Tesla P100 GPU 15Gb. In the indexing step and the quickview retrieval model, we use Elasticsearch\footnote{https://www.elastic.co/} with the configuration setting 8Gb heap size.
Besides, during the experiment with some pre-trained BERT models, the BERT multilingual model produces the best results, so it is used to generate vector representation for the given question and the articles in the dense vector indexing and is used in a supervised model.

\subsection{Quickview Retrieval Result}

Table \ref{tab:wordquickview} shows the results of the word matching method, it is easy to see the superiority in execution time. It only takes 14.43 ms to return the set of 50 candidates and 115.63 ms for 1000 candidates. The results also demonstrate how the title and content of the article have an impact on the retrieval. $Recall@1000$ is only from 0.75 to 0.87 on the datasets if we solely utilize word matching based on either title or content. While using both of them, $Recall@1000$ is nearly 0.9. As a sort of written summary, the title frequently includes important keywords. Consequently, we achieve the best results, 0.9128 in $Recall@1000$, when increasing the question-title matching score by 1.5 times compared to the question content. 

\begin{table}
\caption{$Recall@k$ of Word matching method in quickview retrieval model}
\label{tab:wordquickview}
\centering
\begin{tabular}{lrrrrr}
\hline
\textbf{BM25($\alpha, \beta$)/Top-k} & \textbf{50} & \textbf{100} & \textbf{200} & \textbf{500} & \textbf{1000} \\ \hline
\textbf{Time per Q (ms)} & 14.43 & 20,32 & 31.32 & 63.21& 115,63 \\ \hline
\textbf{Training set} &  &  &  &  &  \\ \hline
BM25(0,1) & 0.6169 & 0.6941 & 0.7586 & 0.8220 & 0.8659 \\ \hline
BM25(1,0) & 0.5674 & 0.6169 & 0.6644 & 0.7172 & 0.7536 \\ \hline
BM25(1,1) & 0.6739 & 0.7478 & 0.8060 & 0.8651 & 0.8998 \\ \hline
BM25(1.5,1) & 0.6942 & 0.7612 & 0.8169 & 0.8740 & 0.9063 \\ \hline
\textbf{Testing set} &  &  &  &  &  \\ \hline
BM25(0,1) & 0.6309 & 0.7103 & 0.7709 & 0.8368 & 0.8747 \\ \hline
BM25(1,0) & 0.5743 & 0.6282 & 0.6792 & 0.7259 & 0.7611 \\ \hline
BM25(1,1) & 0.6943 & 0.7728 & 0.8261 & 0.8798 & 0.9080 \\ \hline
BM25(1.5,1) & \textbf{0.7214} & \textbf{0.7973} & \textbf{0.8453} & \textbf{0.8863} & \textbf{0.9128} \\ \hline
\end{tabular}%
\end{table}

The experimental result of the dense vector matching method is illustrated in Table \ref{tab:densequickview}. Both the dense vector matching on BERT and FastText have lengthy execution times but just average $Recall@k$. In the dense vector indexing method, the articles were indexed at the sentence level, we need to return larger records than the word indexing method based on article level. Calculating the similarity between vectors with large dimensions is also a challenge. Therefore, this method takes a long time to execute. Retrieving 10000 sentences that take 1.7 and 5,2 seconds is not possibly applied in the real-time question-answering system. $R@10000$ is 0.61 for the FastText and 0.67 for the BERT, It is also simple to understand these scores. because the advantage of FastText is a semantic representation at the word level. Whereas BERT is known for its powerful contextual representation of paragraphs, splitting the article into sentences loses this contextual property.

\begin{table}
\caption{$Recall@k$ of quickview retrieval model on the dense vector indexing}
\label{tab:densequickview}
\centering
\begin{tabular}{clcc}
\hline
\textbf{k} & \multicolumn{1}{c}{\textbf{EmbeddingMethod}} & \textbf{R@k} & \textbf{Time(ms)} \\ \hline
\multirow{2}{*}{1000} & FastText(D=300) & 0.40 & 203 \\ \cline{2-4} 
 & BERT(D=768) & 0.38 & 755 \\ \hline
\multirow{2}{*}{2000} & FastText(D=300) & 0.48 & 384 \\ \cline{2-4} 
 & BERT(D=768) & 0.45 & 1,059 \\ \hline
\multirow{2}{*}{5000} & FastText(D=300) & 0.56 & 896 \\ \cline{2-4} 
 & BERT(D=768) & 0.60 & 2,433 \\ \hline
\multirow{2}{*}{10000} & FastText(D=300) & 0.61 & 1,757 \\ \cline{2-4} 
 & BERT(D=768) & 0.67 & 5,204 \\ \hline
\end{tabular}%
\end{table}

Based on the aforementioned experiment results, we decided to build the quickview retrieval model using BM25 with the $\alpha = 1.5$ and $\beta = 1$. For the real-time response, we obtain respectable $Recall@k$ scores of 0.7214, 0.7973 and 0.8453 for the $k$ values in (50, 100, 200), which indicates that the number of candidates will be returned following this phase.

\subsection{End-to-end Question Answering System Result}

Table \ref{tab:end-to-end200} indicates the experimental results of the end-to-end question answering system result with a top 200 candidates from the quickview retrieval model. The word-matching model with BM25 and the supervised model built from the original data gives F2 score is about 0.38. The ensemble model outperforms the other models in F2 score with 0.6007, which is 22\% higher than the single models. As was pointed out in the previous section, lexical and semantic similarity are highly dissimilar. But we believe they can cooperate and support one another. Results certainly support that. Table \ref{tab:end-to-end200}  also clearly illustrates the contribution of the weak label dataset. It improved the supervised machine learning model's F2 score by 8\%. The weak label data continues to have an impact on the F2 score when the lexical and semantic matching models are combined. The ensemble model that used the weak label data had a 1\% increase in F2 scores.  

\begin{table}
\caption{The result of end-to-end QA system result with $top_k=200$}
\label{tab:end-to-end200}
\centering
\begin{tabular}{lccc}
\hline
\textbf{Model}  & \textbf{R} & \textbf{P} & \textbf{F2} \\ \hline
Quickview Model(1.5,1) & 0.4454 & 0.2399 & 0.3803 \\ \hline
Supervised Model (original data) & 0.6165 & 0.1461 & 0.3750 \\ \hline
Supervised Model (full data) & 0.6651 & 0.1998 & 0.4538 \\ \hline
Ensemble Model (original data) & \textbf{0.6681} & 0.4080 & 0.5925 \\ \hline
Ensemble Model (full data) & 0.6651 & \textbf{0.4331} & \textbf{0.6007} \\ \hline
\end{tabular}%
\end{table}

Additionally, there is a sizeable distinction between precision and recall. The recall  is given more consideration because of its great impact on F2 score. We discovered that similarity in lexical and semantics has the same effect during the experimental and evaluation phases. Consequently, $\gamma$ is set at 0.5. Infer time is also a remarkable point in the construction of the question-answering system, which shows the feasibility of the system when applied in practice. 

Table \ref{tab:end-to-end} illustrate the results with  the computational resources in the experimental environment, we can use the model with the top 50|100 candidates with an execution time of 1 second and 1.7 seconds per question. Their F2 scores are also only 2-5\% lower than the best model.

\begin{table}
\caption{The result of end-to-end QA system result with ensemble model}
\label{tab:end-to-end}
\centering
\begin{tabular}{lcccc}
\hline
\textbf{Ensemble Model} & \textbf{R} & \textbf{P} & \textbf{F2} & \textbf{Time(s)} \\ \hline
(full data, k=20) & 0.5677 & 0.4034 & 0.5252 & 0.5 \\ \hline
(full data, k=50) & 0.5842 & 0.4428 & 0.5491 & 1 \\ \hline
(full data, k=100) & 0.6222 & \textbf{0.4475} & 0.5771 & 1.7 \\ \hline
(full data, k=200) & 0.6651 & 0.4331 & \textbf{0.6007} & 3.4 \\ \hline
(full data, k=500) & \textbf{0.6793} & 0.4015 & 0.5967 & 8.5 \\ \hline
(full data, k=1000) & 0.6583 & 0.4261 & 0.5936 & 17 \\ \hline
\end{tabular}%
\end{table}

Table \ref{tab:compare} shows that our recall and F2 scores are incredibly high when compared to the Attentive CNN \cite{kien2020answering} and the Paraformer \cite{nguyen2022attentive} models (0.6651 and 0.6007). Their models return small amounts of related articles, while our system is designed to return flexible amounts of articles with $threshold$. This explains why their precision is great, about 0.5987, whereas our precision is only 0.4331. A set of thresholds for each $top-k$ is listed in Table \ref{tab:threshold}.

\begin{table}
\caption{The result compared with other research groups}
\label{tab:compare}
\centering
\begin{tabular}{lccc}
\hline
\textbf{Systems} & \textbf{R} & \textbf{P} & \textbf{F2} \\ \hline
Attentive CNN \cite{kien2020answering} & 0.4660 & 0.5919 & 0.4774 \\ 
Paraformer \cite{nguyen2022attentive} & 0.4769 & \textbf{0.5987} & 0.4882 \\ \hline
Our model (k=50) & 0.5842 & 0.4428 & 0.5491 \\ 
Our model (k=100) & 0.6222 & 0.4475 & 0.5771 \\ 
Our model (k=200) & \textbf{0.6651} & 0.4331 & \textbf{0.6007} \\ \hline
\end{tabular}%
\end{table}

\begin{table}
\caption{Threshold list of the ensemble model}
\centering
\label{tab:threshold}
\begin{tabular}{lcccccc}
\hline
\textbf{$top\_k$} & 20 & 50 & 100 & 200 & 500 & 1000 \\ \hline
\textbf{$threshold$} & 0.38 & 0.28 & 0.26 & 0.26 & 0.25 & 0.2 \\ \hline
\end{tabular}%
\end{table}

Table \ref{tab:error} describes an example of our legal question-answering system, compared with Paraformer~\cite{nguyen2022attentive}. A small number of related articles are frequently returned by Paraformer models. Our system is more flexible with 3 returned related articles. While the gold label number is 2. As an outcome, a paragraph model like Paraformer is produced that has great precision but low recall, whereas our method leans in the opposite direction. Since recall has a greater impact on F2 scores, our model has a significantly higher F2 score of 11\%.

\begin{table}
\caption{An output example of ours System, compared with Paraformer \cite{nguyen2022attentive}.}
\label{tab:error}
\selectlanguage{vietnamese}
\fontsize{6.5pt}{7.5pt}\selectfont
\centering
\begin{tabular}{|p{0.78\textwidth}|p{0.06\textwidth}|p{0.07\textwidth}|p{0.05\textwidth}|}
\hline
\textbf{Question:} Vay tiền để kinh doanh nhưng không còn khả năng chi trả phải trả lãi suất thì như thế nào?& Ours & Para-former & Gold \\
\textit{(In the case of insolvency, how does one address the issue of paying the interest on a business loan?)} &  &   &  \\ \hline
 \textbf{Candidate 1:} Id: Article 357 from Doc 91/2015/QH13 & 1 & 1 & 1 \\ 
 \textbf{Title:} Trách nhiệm do chậm thực hiện nghĩa vụ trả tiền & & & \\
\textit{(Liability for late performance of the obligation to pay)} &  &  &  \\
 \textbf{Content:} 1. Trường hợp bên có nghĩa vụ chậm trả tiền thì bên đó phải trả lãi đối với số tiền chậm trả tương ứng với thời gian chậm trả. & & & \\
 2. Lãi suất phát sinh do chậm trả tiền được xác định theo thỏa thuận của các bên nhưng không được vượt quá mức lãi suất được quy định tại khoản 1 Điều 468; nếu không có thỏa thuận thì thực hiện theo quy định tại khoản 2 Điều 468. &  &  &  \\ 
 \textit{(1. Where the obligor makes late payment, then it must pay interest on the unpaid amount
corresponding to the late period.} & & & \\
 \textit{2. Interest arising from late payments shall be determined by agreement of the parties, but may not exceed the interest rate specified in paragraph 1 of Article 468 of this Code; if there no agreement mentioned above, the Clause 2 of Article 468 of this Code shall apply.)} &  &  &  \\ \hline
 \textbf{Candidate 2:} Id: Article 466 from Doc 91/2015/QH13 & 1 & 0 & 0 \\
 \textbf{Title:} Nghĩa vụ trả nợ của bên vay & & & \\
 \textit{(Obligations of borrowers to repay loans)} &  &  &  \\ 
\textbf{Content:} {[}...{]}5. Trường hợp vay có lãi mà khi đến hạn bên vay không trả hoặc trả không đầy đủ thì bên vay phải trả lãi như sau: & & & \\
 a) Lãi trên nợ gốc theo lãi suất thỏa thuận trong hợp đồng tương ứng với thời hạn vay mà đến hạn chưa trả; trường hợp chậm trả thì còn phải trả lãi theo mức lãi suất quy định tại khoản 2 Điều 468 của Bộ luật này; &  &  &  \\ 
 b) Lãi trên nợ gốc quá hạn chưa trả bằng 150\% lãi suất vay theo hợp đồng tương ứng với thời gian chậm trả, trừ trường hợp có thỏa thuận khác. &  &  &  \\ 
 \textit{({[}...{]} 5. If a borrower fails to repay, in whole or in part, a loan with interest, the borrower must pay:} & & & \\
 \textit{a) Interest on the principal as agreed in proportion to the overdue loan term and interest at the rate prescribed in Clause 2 Article 468 in case of late payment;} &  &  &  \\ 
 \textit{b) Overdue interest on the principal equals one hundred and fifty (150) per cent of the interest rate in proportion to the late payment period, unless otherwise agreed.)} &  &  &  \\ \hline
\textbf{Candidate 3:} Id: Article 468 from Doc 91/2015/QH13 & 1 & 0 & 1 \\ 
 \textbf{Title:} Lãi suất \textit{(Interest rates)} & & & \\
\textbf{Content:} 1. Lãi suất vay do các bên thỏa thuận.{[}...{]} &  &  &  \\ 
 2. Trường hợp các bên có thỏa thuận về việc trả lãi, nhưng không xác định rõ lãi suất và có tranh chấp về lãi suất thì lãi suất được xác định bằng 50\% mức lãi suất giới hạn quy định tại khoản 1 Điều này tại thời điểm trả nợ. &  &  &  \\
 \textit{(1. The rate of interest for a loan shall be as agreed by the parties.[...]} & & & \\
 \textit{2. Where parties agree that interest will be payable but fail to specify the interest rate, or where there is a dispute as to the interest rate, the interest rate for the duration of the loan shall equal 50\% of the maximum interest prescribed in Clause 1 of this Article at the repayment time.)} &  &  &  \\ \hline
\end{tabular}%
\end{table}


Our model predicts that ``Article 466 from Doc 91/2015/QH13'' is relevant to the given query but the gold label is 0. Considering this article, we believe the article is pertinent to the given question but it seems that the annotator's point of view is different. In addition, we discovered some similar cases in our error analysis. 
Defining and agreeing on a measure of relevance is an important research question that needs the participation of the AI and Law community in its research.
This not only benefits the development of automated methods but also makes legal judgments and decisions more reliable and accurate.

\section{Conclusions}
In this paper, we present a method to improve performance in the task of legal question answering for Vietnamese using language models through weak labeling. By demonstrating the effectiveness of this method through experiments, we verify the hypothesis that improving the quality and quantity of datasets is the right approach for this problem, especially in low-resource languages like Vietnamese. The results of our work can provide valuable insights and serve as a reference for future attempts to tackle similar challenges in low-resource legal question-answering.

\section*{Acknowledgement}
This work was supported by VNU University of Engineering and Technology under project number CN22.09.

\bibliographystyle{splncs04}
\bibliography{references.bib}

\begin{thebibliography}{10}
\providecommand{\url}[1]{\texttt{#1}}
\providecommand{\urlprefix}{URL }
\providecommand{\doi}[1]{https://doi.org/#1}

\bibitem{brown2020language}
Brown, T., Mann, B., Ryder, N., Subbiah, M., Kaplan, J.D., Dhariwal, P.,
  Neelakantan, A., Shyam, P., Sastry, G., Askell, A., et~al.: Language models
  are few-shot learners. Advances in neural information processing systems
  \textbf{33},  1877--1901 (2020)

\bibitem{devlin2018bert}
Devlin, J., Chang, M.W., Lee, K., Toutanova, K.: {BERT}: Pre-training of deep
  bidirectional transformers for language understanding. In: NAACL. pp.
  4171--4186 (Jun 2019)

\bibitem{feldman2019multi}
Feldman, Y., El-Yaniv, R.: Multi-hop paragraph retrieval for open-domain
  question answering. In: Proceedings of the 57th Annual Meeting of the
  Association for Computational Linguistics. pp. 2296--2309 (2019)

\bibitem{hu2014convolutional}
Hu, B., Lu, Z., Li, H., Chen, Q.: Convolutional neural network architectures
  for matching natural language sentences. In: Advances in neural information
  processing systems. pp. 2042--2050 (2014)

\bibitem{huang2013learning}
Huang, P.S., He, X., Gao, J., Deng, L., Acero, A., Heck, L.: Learning deep
  structured semantic models for web search using clickthrough data. In:
  Proceedings of the 22nd ACM international conference on Information \&
  Knowledge Management. pp. 2333--2338 (2013)

\bibitem{iyyer2014neural}
Iyyer, M., Boyd-Graber, J., Claudino, L., Socher, R., Daum{\'e}~III, H.: A
  neural network for factoid question answering over paragraphs. In:
  Proceedings of the 2014 conference on empirical methods in natural language
  processing (EMNLP). pp. 633--644 (2014)

\bibitem{joulin2017bag}
Joulin, A., Grave, E., Bojanowski, P., Mikolov, T.: Bag of tricks for efficient
  text classification. In: Proceedings of the 15th Conference of the European
  Chapter of the Association for Computational Linguistics: Volume 2, Short
  Papers. pp. 427--431. Association for Computational Linguistics (April 2017)

\bibitem{kien2020answering}
Kien, P.M., Nguyen, H.T., Bach, N.X., Tran, V., Le~Nguyen, M., Phuong, T.M.:
  Answering legal questions by learning neural attentive text representation.
  In: Proceedings of the 28th International Conference on Computational
  Linguistics. pp. 988--998 (2020)

\bibitem{adam}
Kingma, D., Ba, J.: Adam: A method for stochastic optimization. International
  Conference on Learning Representations  (12 2014)

\bibitem{nguyen2022attentive}
Nguyen, H.T., Phi, M.K., Ngo, X.B., Tran, V., Nguyen, L.M., Tu, M.P.: Attentive
  deep neural networks for legal document retrieval. Artificial Intelligence
  and Law pp. 1--30 (2022)

\bibitem{palangi2016deep}
Palangi, H., Deng, L., Shen, Y., Gao, J., He, X., Chen, J., Song, X., Ward, R.:
  Deep sentence embedding using long short-term memory networks: Analysis and
  application to information retrieval. IEEE/ACM Transactions on Audio, Speech,
  and Language Processing  \textbf{24}(4),  694--707 (2016)

\bibitem{rabelo2022overview}
Rabelo, J., Goebel, R., Kim, M.Y., Kano, Y., Yoshioka, M., Satoh, K.: Overview
  and discussion of the competition on legal information extraction/entailment
  (coliee) 2021. The Review of Socionetwork Strategies  \textbf{16}(1),
  111--133 (2022)

\bibitem{reimers2019sentence}
Reimers, N., Gurevych, I.: Sentence-bert: Sentence embeddings using siamese
  bert-networks. In: Proceedings of the 2019 Conference on Empirical Methods in
  Natural Language Processing and the 9th International Joint Conference on
  Natural Language Processing (EMNLP-IJCNLP). pp. 3982--3992 (2019)

\bibitem{robertson1994some}
Robertson, S.E., Walker, S.: Some simple effective approximations to the
  2-poisson model for probabilistic weighted retrieval. In: SIGIR’94. pp.
  232--241. Springer (1994)

\bibitem{sanagavarapu2022disentangling}
Sanagavarapu, K., Singaraju, J., Kakileti, A., Kaza, A., Mathews, A., Li, H.,
  Brito, N., Blanco, E.: Disentangling indirect answers to yes-no questions in
  real conversations. In: Proceedings of the 2022 Conference of the North
  American Chapter of the Association for Computational Linguistics: Human
  Language Technologies. pp. 4677--4695 (2022)

\bibitem{sugathadasa2019legal}
Sugathadasa, K., Ayesha, B., de~Silva, N., Perera, A.S., Jayawardana, V.,
  Lakmal, D., Perera, M.: Legal document retrieval using document vector
  embeddings and deep learning. In: Intelligent Computing: Proceedings of the
  2018 Computing Conference, Volume 2. pp. 160--175. Springer (2019)

\bibitem{tai-etal-2015-improved}
Tai, K.S., Socher, R., Manning, C.D.: Improved semantic representations from
  tree-structured long short-term memory networks. In: Proceedings of the 53rd
  Annual Meeting of the Association for Computational Linguistics and the 7th
  International Joint Conference on Natural Language Processing (Volume 1: Long
  Papers). pp. 1556--1566. Association for Computational Linguistics, Beijing,
  China (Jul 2015)

\bibitem{tran2020encoded}
Tran, V., Le~Nguyen, M., Tojo, S., Satoh, K.: Encoded summarization:
  summarizing documents into continuous vector space for legal case retrieval.
  Artificial Intelligence and Law  \textbf{28},  441--467 (2020)

\bibitem{vaswani2017attention}
Vaswani, A., Shazeer, N., Parmar, N., Uszkoreit, J., Jones, L., Gomez, A.N.,
  Kaiser, {\L}., Polosukhin, I.: Attention is all you need. Advances in neural
  information processing systems  \textbf{30} (2017)

\bibitem{vuong2022sm}
Vuong, Y.T.H., Bui, Q.M., Nguyen, H.T., Nguyen, T.T.T., Tran, V., Phan, X.H.,
  Satoh, K., Nguyen, L.M.: Sm-bert-cr: a deep learning approach for case law
  retrieval with supporting model. Artificial Intelligence and Law pp. 1--28
  (2022)

\end{thebibliography}
\end{document}